\documentclass[sigconf]{acmart}
\AtBeginDocument{%
  \providecommand\BibTeX{{%
    \normalfont B\kern-0.5em{\scshape i\kern-0.25em b}\kern-0.8em\TeX}}}

\setcopyright{acmcopyright}
\copyrightyear{2021}
\acmYear{2021}

\acmConference[CIKM '22]{CIKM '22}{June 03--05, 2018}{Arizona}
\acmBooktitle{CIKM '22,
  February 21--25, 2022, Arizona}

\usepackage[utf8]{inputenc} 
\usepackage[T1]{fontenc}    
\usepackage{hyperref}       
\usepackage{url}            
\usepackage{amsfonts}       
\usepackage{microtype}      
\usepackage{multirow} 
\usepackage{subcaption}
\usepackage{xcolor}         

\newcommand{\tabincell}[2]{\begin{tabular}{@{}#1@{}}#2\end{tabular}}

\begin{document}

\title{Evaluating Modules in Graph Contrastive Learning}


\author{Ganqu Cui$^{1,2}$\footnotemark[1], Yufeng Du$^{1,2}$\footnotemark[1], Cheng Yang$^{4}$\footnotemark[2], Jie Zhou$^{1,2}$, Liang Xu$^{1,2}$,\\ Xing Zhou$^{6}$, Xingyi Cheng$^{6}$, Zhiyuan Liu$^{1,2,3,4}$\footnotemark[2]}

\affiliation{%
  \institution{$^1$Dept. of Comp. Sci. \& Tech., Institute for AI, Tsinghua University, Beijing}
  \institution{$^2$Beijing National Research Center for Information Science and Technology}
  \institution{$^3$Institute Guo Qiang, Tsinghua University, Beijing}
  \institution{$^{4}$International Innovation Center of Tsinghua University, Shanghai}
  \institution{$^5$School of Computer Science, Beijing University of Posts and Telecommunications $^6$Tencent Inc.}
  \country{China}\\
  \{cgq19,duyf18\}@mails.tsinghua.edu.cn
}

\renewcommand{\shortauthors}{Cui and Du, et al.}
\begin{abstract}
The recent emergence of contrastive learning approaches facilitates the application on graph representation learning (GRL), introducing graph contrastive learning (GCL) into the literature. These methods contrast semantically similar and dissimilar sample pairs to encode the semantics into node or graph embeddings. However, most existing works only performed \textbf{model-level} evaluation, and did not explore the combination space of modules for more comprehensive and systematic studies.
For effective \textbf{module-level} evaluation, we propose a framework that decomposes GCL models into four modules: (1) a \textbf{sampler} to generate anchor, positive and negative data samples (nodes or graphs); (2) an \textbf{encoder} and a \textbf{readout} function to get sample embeddings; (3) a \textbf{discriminator} to score each sample pair (anchor-positive and anchor-negative); and (4) an \textbf{estimator} to define the loss function. Based on this framework, we conduct controlled experiments over a wide range of architectural designs and hyperparameter settings on node and graph classification tasks. 
Specifically, we manage to quantify the impact of a single module, investigate the interaction between modules, and compare the overall performance with current model architectures.
   Our key findings include a set of module-level guidelines for GCL, e.g., simple samplers from LINE and DeepWalk are strong and robust; an MLP encoder associated with Sum readout could achieve competitive performance on graph classification. 
Finally, we release our implementations and results as OpenGCL, a modularized toolkit that allows convenient reproduction, standard model and module evaluation, and easy extension. OpenGCL is available at \url{https://github.com/thunlp/OpenGCL}.
\renewcommand{\thefootnote}{\fnsymbol{footnote}} 
\footnotetext[1]{Both authors contributed equally to this research.}
\footnotetext[2]{Corresponding authors.}
\end{abstract}



\maketitle

\section{Introduction}
Learning representations with better utilization of unlabelled data has been a crucial issue in deep learning.
Among such approaches, self-supervised learning designs heuristic pretext tasks to learn representations with self-generated labels~\cite{liu2020self}.
Recent advances in self-supervised learning mainly concentrate on \textit{contrastive learning}, which discriminates semantically similar and dissimilar sample pairs following the InfoMax principle~\cite{linsker1988self}. Sustainable progress has been made in domains such as computer vision~\cite{chen2020simple, he2020momentum, oord2018representation} and natural language processing~\cite{devlin2019bert, kong2020mutual, mikolov2013distributed}.

In the literature of graph representation learning (GRL), traditional models directly train node embeddings by graph structure. DeepWalk~\cite{perozzi2014deepwalk} assumes that concurrences of nodes in random walks can reveal structural similarity. LINE~\cite{tang2015line} presumes the first-order and second-order similarities of neighboring nodes. 
With the success of graph neural networks (GNNs)~\cite{kipf2016semi}, graph auto-encoder (GAE)~\cite{kipf2016variational} combines feature and structure information, and makes a significant advance. Recently, there also emerge several contrastive learning models~\cite{hassani2020contrastive, sun2020infograph, velickovic2019deep, you2020graph, zhu2021graph} which contrast between different graph views. These methods achieve state-of-the-art performance and even outperform some semi-supervised models~\cite{hassani2020contrastive, zhu2021graph}.
However, despite the potential of self-supervised GRL, current \textit{analysis} and \textit{evaluation} are limited to \textbf{model level}. In the following paragraphs, we will demonstrate this issue and stress the importance of \textbf{module-level} analysis and evaluation.

 
Firstly, current self-supervised GRL models are developed independently. They mainly concentrate on designing new pretext tasks and their architectural differences and similarities are barely explored. For example, the training objectives of LINE-2nd~\cite{tang2015line}  and GAE~\cite{kipf2016variational} both presume that neighboring nodes are analogous and optimize the inner product similarity (see appendix for details). As we will discuss later, both models can fit into the contrastive learning framework, with a mere difference in graph encoder. 
However, a rough model-level analysis will neglect these internal model architectures.
Therefore, to get better understandings of graph contrastive learning, we propose a modularized framework, and show how eight classical and state-of-the-art self-supervised GRL models can be fit into the framework.
Our framework interprets graph constrastive learning with 4 disjoint modules: (1) The context sampler, which chooses anchor, positive, and negative data samples; (2) the graph encoder and readout function, which project samples to low-dimensional representations; (3) the discriminator, which assigns a score to each anchor-context pair; (4) the estimator, which provides a contrastive loss.

Secondly, model-level evaluation is also insufficient for a thorough analysis. As the models are specific combinations of disjoint modules, module-level evaluation allows us to gain more insights about how to learn good representations by effective architectural design.
In this work, we conduct thorough evaluations over various modules by controlled experiments. We manage to investigate the impact of each single module, explore the pairwise combination effect of modules, and compare our best module combinations with state-of-the-art models.
Our key results include: (1) simple LINE and DeepWalk samplers are surprisingly powerful and robust; (2) the combination of Sum readout function and MLP encoder are comparable with GNN-based architectures on graph classification; (3) the best model architectures from our framework can achieve state-of-the-art performances, which illustrates the potential of the framework.

In addition, most graph contrastive learning model implementations are not fully modularized as well, making it difficult for module-level evaluation. Thus, a modularized and extendable toolkit is also needed. To deal with this issue, we develop OpenGCL, an open-source toolkit for the implementation and evaluation of modularized graph contrastive learning model.

We sum up our contributions as follows: 
\begin{enumerate}
    \item \textbf{Framework.} We propose a modularized framework for GCL and analyze the distinctions and similarities of several typical models under the framework.
    \item \textbf{Evaluation.} We conduct extensive experiments on a wide range of architectural designs, hyperparameter settings, and benchmarks. The experimental results and analysis help us explore the key factors in learning good representations.
    \item \textbf{Toolkit.} We release OpenGCL, an open-source toolkit for graph contrastive learning. OpenGCL allows modularized implementation and evaluation of various models, encouraging further research on GCL.
\end{enumerate}

\section{Related Work}
\subsection{Contrastive Learning}
For contrastive learning, the key issue is the augmentation strategy to generate similar pairs. In the computer vision domain, additional views are usually produced by randomly cropping or flipping the original image~\cite{chen2020simple,oord2018representation}, by which most semantics of images are preserved. 

Graphs, however, usually have latent semantics, which brings challenges when designing contrastive learning models.
Early methods mainly focused on node-level contrast, which managed to find similar context nodes for a given anchor node~\cite{kipf2016variational, perozzi2014deepwalk, tang2015line}. The context nodes are selected locally from neighborhoods or random walks and the contrastive learning principle is entailed implicitly. Motivated by Deep InfoMax~\cite{hjelm2019learning}, some methods~\cite{hassani2020contrastive, sun2020infograph, velickovic2019deep} proposed to contrast local and global representations for learning multi-scale graph structure representations. Most recently, models with newly designed augmentation strategies like node dropping and attribute masking were proposed~\cite{you2020graph, zhu2021graph}.

\subsection{Frameworks of GRL Models}
The idea of unifying GRL models has been studied from various perspectives~\cite{hamilton2017representation, qiu2018network, you2020graph}. Conventional shallow embedding models can be summarized by matrix factorization~\cite{qiu2018network,yang2017fast}. However, the matrix factorization framework cannot easily incorporate feature information. Therefore, a more comprehensive encoder-decoder framework~\cite{hamilton2017representation} was introduced to interpret algorithms based on matrix factorization, random walk and graph neural networks. 

In terms of contrastive learning, You et al.~\cite{you2020graph} proposed a contrastive framework for graph-level representation learning, which mainly focused on the design of different pretext tasks. Some recent surveys~\cite{liu2021graph, wu2021self, xie2021self} also summarized graph contrastive learning models, but they did not compare the effect of different modules by experiments. Recently, there is a concurrent work~\cite{zhu2021empirical} which also discussed and evaluated graph contrastive learning.
Compared with this work, we conduct controlled random experiments to comprehensively evaluate modules. Moreover, we investigate the pairwise combination effect and overall performance, which makes our work more systematic.

\subsection{Evaluation of GRL Models}
Multiple research works concentrated on evaluating various GRL models~\cite{dwivedi2020benchmarking, errica2020fair, mesquita2020rethink, shchur2018pitfalls, you2020design}. Shchur et al.~\cite{shchur2018pitfalls} explored the impact of dataset split for GNN models and concluded that different dataset splits lead to dramatically different model performances. Some works~\cite{dwivedi2020benchmarking, hu2020open} further proposed new multi-scale benchmarks for appropriate GNN evaluation. On graph classification task, Errica et al.~\cite{errica2020fair} provided a rigorous comparison for graph classification. Mesquita et al.~\cite{mesquita2020rethink} challenged all the local pooling algorithms and demonstrated that local graph pooling is not crucial for learning graph representations.
However, the aforementioned research works are limited to model-level evaluation of existing models. We break down full models into modules, which helps us find out what plays a bigger role inside a model. You et al.~\cite{you2020design} discussed the architectural designs of GNN models, such as batch size and the activation function. By abundant experiments, they provided comprehensive guidelines for GNN designs over various tasks. 
In contrast, we focus on module-level architectures rather than network designs.

\section{Preliminaries}

Given a graph $\mathcal{G}=(\mathcal{V}, \mathcal{E})$, where $\mathcal{V}$ is the vertex set and $\mathcal{E}$ is the edge set, the graph structure can be represented by an adjacency matrix $\mathbf{A}$. In some cases, nodes in graph $\mathcal{G}$ are associated with features $\mathbf{X}$. Graph representation learning aims to map graph substructures (nodes, edges, subgraphs etc.) to low-dimensional embeddings.

 After we get the node representations, we can summarize the graph-level representations by a readout function.
The learned representations are often used in downstream graph learning tasks. In this paper, we focus on node classification and graph classification tasks.

\section{Proposed Framework}
 We propose a modularized framework for graph contrastive learning, which consists of four modules: 
\begin{enumerate}
    \item \textbf{Context sampler.} Following the priors inside  data distributions, a context sampler provides an augmentation mechanism on graph data, which generates semantically similar (positive) and dissimilar (negative) contexts for the given anchor data point. For each batch, the context sampler chooses one anchor $x$, one positive context $k^+$ and $N$ negative contexts $\{k^-_i\}_{i=1}^N$.
    \item \textbf{Graph encoder.} The encoder $\texttt{Enc}$ maps selected data points to distributional vectors. More specifically, we learn node embeddings by node encoders, and get graph embeddings by a following readout function.
    \item \textbf{Discriminator.} The discriminator $\texttt{Disc}$ scores each anchor-context pair in the batch according to the agreement between them. The score function $f$ is defined as
\begin{equation}
f(x,k)=\texttt{Disc}(\texttt{Enc}(x), \texttt{Enc}(k)).
\end{equation}
    \item \textbf{Mutual information estimator.} The mutual information estimator $I_\texttt{Est}$ defines the loss function, which enforces the representations of positive pairs (anchor and positive contexts) to be close, and those of negative pairs (anchor and negative contexts) to be orthogonal. We define the optimization objective as follows:
    \begin{equation}
    \max\mathbb{E}_{x, k^+, \{k^-_i\}_{i=1}^N}\left[ I_\texttt{Est}\left(f(x,k^+); \left\{f(x,k^-_i)\right\}_{i=1}^N\right)\right].
    \end{equation}
\end{enumerate}

\begin{figure*}[ht]
    \centering
    \includegraphics[width=\linewidth]{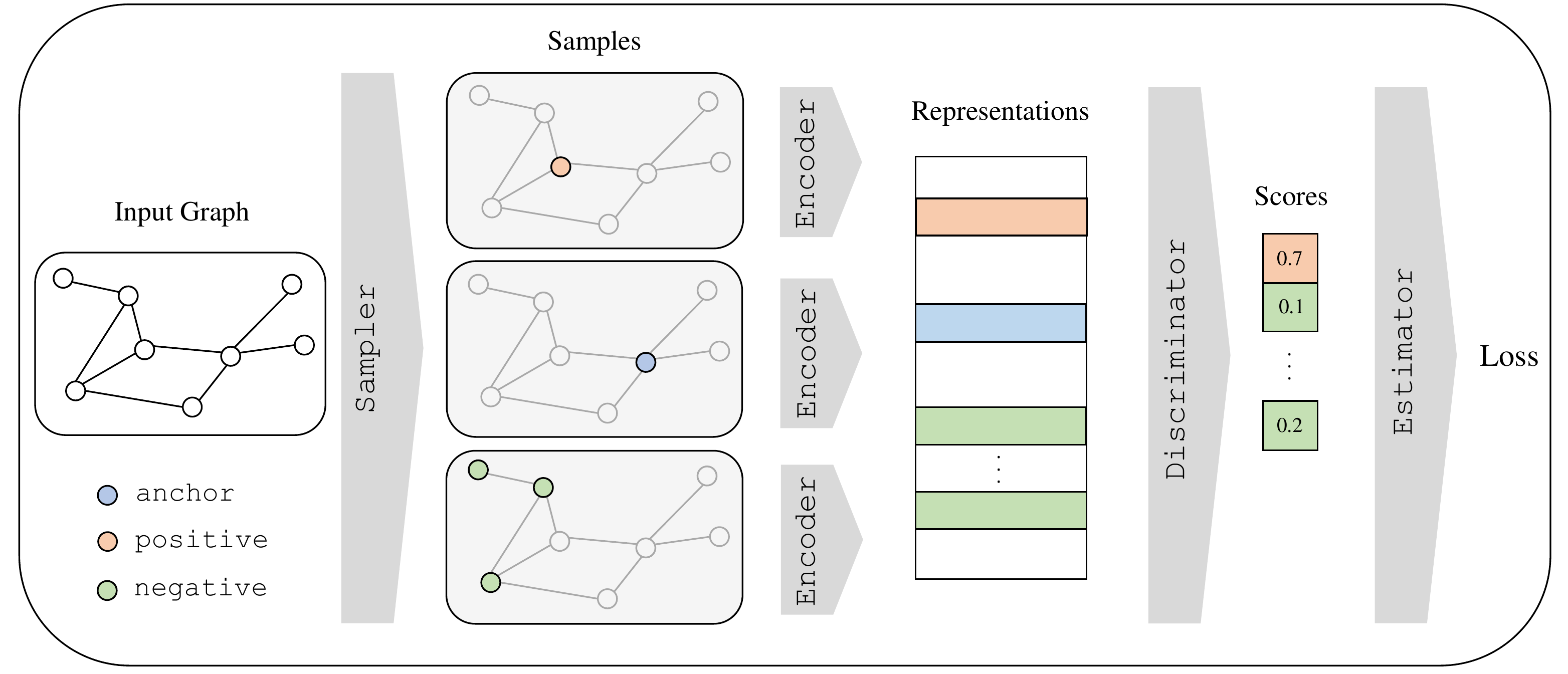}
    \caption{Our framework of GCL. (1) The anchor, positive and negative data samples are selected by the sampler following an underlying prior data distribution. (2) A shared graph encoder is then applied to project data samples to low-dimensional representations. (3) The discriminator scores anchor-context pairs. (4) The estimator brings positive pairs closer and separates negative pairs apart by a contrastive loss.}
    \label{fig:model}
\end{figure*}

Next we give a detailed discussion over possible instantiations of each module.

\subsection{Sampler}
 One major distinction in different sampling strategies lies in the ``views'' of data. Samplers in LINE~\cite{tang2015line}, GAE~\cite{kipf2016variational} and DeepWalk~\cite{perozzi2014deepwalk} focus on the same local view of graph structure, where samples come from nodes in the original graph. Samplers in DGI~\cite{velickovic2019deep} and InfoGraph~\cite{sun2020infograph} adapt the idea of Deep InfoMax~\cite{hjelm2019learning} to graph domain; they maximize the mutual information between global and local views in the original dataset by contrasting nodes against graphs. More recently, MVGRL~\cite{hassani2020contrastive}, GCA~\cite{zhu2021graph} and GraphCL~\cite{you2020graph} develop samplers that generate views by augmenting the given graphs; the corresponding nodes and/or graphs in different views are thought to be similar. 
Out of the eight mentioned models, we conclude six samplers based on their views and sampling processes:
\begin{enumerate}
    \item Sampler in LINE (and GAE) selects neighboring nodes as anchors and positive samples. Negative samples are random nodes in the dataset.
    \item Sampler in DeepWalk performs random walk and applies Skip-gram model to the resulting paths, where anchor-positive pairs are taken by sliding windows. Negative samples are random nodes in the dataset.
    \item Sampler in DGI (and InfoGraph) selects graphs as anchors. Nodes are seen as positive samples if they are from the anchor graph, and as negative samples if not. In datasets with one single graph, negative nodes are chosen from a ``negative graph'' which is generated by shuffling node features.
    \item Sampler in MVGRL is similar to that in DGI, except that the positive and negative samples are nodes from diffusion graphs like personalized PageRank~\cite{page1999pagerank}.
    \item Sampler in GCA takes up edge dropping and attribute masking as augmentation strategies and contrasts nodes between graph augmentations $G_A$ and $G_B$ ($A$ and $B$ are randomly selected augmentation strategies). It takes one node in $G_A$ as anchor, the corresponding node in $G_B$ as positive sample, and all the other nodes from both graphs as negative samples. 
    \item Sampler in GraphCL contrasts global augmentative views. It adapts edge dropping, attribute masking, node dropping and subgraph sampling for augmentation.  Augmentations $G_A$ and $G_B$ from one graph are selected as anchor and positive sample, while augmentation $G^\prime_B$ from another graph is the negative sample.
\end{enumerate}

\subsection{Encoder} 
\subsubsection{Node Encoder}
A node encoder, which converts nodes to embeddings, is the first part of a encoder. Conventional GRL models (DeepWalk, LINE) usually learn node embeddings directly, thus the encoders are simply lookup tables. However, lookup tables initialize node embeddings randomly and independently, which can not utilize node feature and graph topology structure. With the proposal of GCN, many recent models (GAE, DGI) adopt GCN as their encoders to capture neighborhood dependency and node feature information. Some variants of GCN are also widely adopted. GAT~\cite{velickovic2018graph} uses the attention mechanism to better model the relative influence between node pairs. Graph isomorphism network (GIN)~\cite{xu2019how} is proved to be as powerful as the Weisfeiler-Lehman graph isomorphism test thus is more expressive than GCN. Instead of average aggregator, GIN sums over neighborhood hidden representations to update the target node's hidden representation. GIN could capture the size information of graphs.
Therefore, InfoGraph and GraphCL alter GCN with GIN as it better suits graph-level tasks. 

\subsubsection{Readout Function}
Additionally, for models that need to learn graph-level representations, a readout function is required to summarize node representations. Two basic readout functions are average and sum functions. DGI takes the average of node embeddings to get graph embeddings. InfoGraph and GraphCL sum over node embeddings to encode graph size information. To utilize the mid-layer node representations, following JK-Net~\cite{xu2018representation}, MVGRL concatenates the summation of node representation in each encoder layer for graph representations.

\subsection{Discriminator} The discriminator measures the similarity of node or graph pairs. Intuitively, inner product is the most straightforward score function, which is selected by most models. DGI adopts a bilinear discriminator instead to score between local and global (node and graph) representations. 

\subsection{Estimator} Several mutual information estimators have been introduced in previous works. Based on Jensen-Shannon divergence~\cite{nowozin2016f}, the objective functions in DeepWalk, LINE, GAE, DGI, InfoGraph and MVGRL are binary cross-entropy over positive and negative scores:
\begin{equation}
    I_{\texttt{JSD}} = \sigma(f(x,k^+)) + \frac{1}{N}\sum_{i=1}^{N}\sigma\left(-f(x,k^-_{i})\right),
\end{equation}
where $\sigma$ is the LogSigmoid function, which is defined as 
\begin{equation}
    \text{LogSigmoid}(x)=\log\frac{1}{1+\exp(-x)}
\end{equation}

Based on Noise-Contrastive Estimation of KL-divergence, InfoNCE estimator is also commonly used in contrastive learning~\cite{oord2018representation}, which is defined as
\begin{equation}
    I_{\texttt{InfoNCE}} = \log\frac{\exp{\left(f(x,k^+)\right)}}{\exp{\left(f(x,k^+)\right)}+\sum_{i=1}^{N}\exp{\left(f(x,k^-_i)\right)}}.
\end{equation}
GCA and GraphCL use InfoNCE estimator. Note that InfoNCE estimator usually adopts in-batch negative samples, where all samples except positive ones in the same mini-batch are selected as negative samples.

\begin{table*}[ht]
    \renewcommand\arraystretch{1}
    \centering
    \caption{Summary of algorithms in our framework.}
    \begin{tabular}{l|cc|c|c|c}
    \toprule
         \begin{tabular}{@{}c@{}} \textbf{Model} \end{tabular}  &  \multicolumn{2}{|c|}{\begin{tabular}{@{}cc@{}}
                                                                \multicolumn{2}{c}{\textbf{Encoder}}\\
                                                                \midrule
                                                                \textbf{Node Encoder} & \textbf{Readout Function}
                                                             \end{tabular}}  & 
                                \begin{tabular}{@{}c@{}} \textbf{Sampler} 
                                \end{tabular} & \begin{tabular}{@{}c@{}} \textbf{Discriminator} \end{tabular} & \begin{tabular}{@{}c@{}} \textbf{Estimator} \end{tabular}\\
    \midrule
       DeepWalk~\cite{perozzi2014deepwalk} & Lookup & - & Random walk & Inner product & JSD \\
       LINE~\cite{tang2015line} & Lookup & - & Neighboring nodes & Inner product &  JSD\\
       GAE~\cite{kipf2016variational} &  GCN & - & Neighboring nodes & Inner product & JSD\\
       DGI~\cite{velickovic2019deep} & GCN & Mean & Local-global views  & Bilinear& JSD\\
       MVGRL~\cite{hassani2020contrastive} & GCN & JK-Net & Diffusion & Inner product & JSD\\
       InfoGraph~\cite{sun2020infograph} & GIN & Sum & Local-global views & Inner product & JSD\\
       GraphCL~\cite{you2020graph} & GIN & Sum & Augmentative global & Inner product & InfoNCE\\
       GCA~\cite{zhu2021graph} & GCN & - & Augmentative local & Inner product & InfoNCE\\ 
     \bottomrule
    \end{tabular}
    \label{tab:framework}
\end{table*}
 
We summarize how several influential algorithms consistently fit into our framework. These models differ in the choice of encoders, samplers, discriminators and estimators. As shown in Table~\ref{tab:framework}, our framework provides an alternative view for self-supervised GRL. In experiments, we factorize these methods and control modules to find out the key issues for learning high-quality representations.

\section{OpenGCL}
To foster our experiments and further researches, we develop an open-source toolkit OpenGCL. OpenGCL is specifically designed for graph contrastive learning with several important features: 

\begin{enumerate}
    \item \textbf{Modularization.} We implement OpenGCL with factorized modules following our framework, which allows users to create any module combination within one command line. Users can reproduce existing models and easily choose the best combinations for specific tasks and datasets.
    \item \textbf{Standardized evaluation.} We enable convenient model-level and module-level evaluation on multiple tasks and datasets. With simple scripts, users can generate, run and manage experiments.
    \item \textbf{Extensibility.} Users may easily implement their modules, such as new samplers or graph pooling readout functions. New tasks and datasets can also be imported with ease.
\end{enumerate}

All of the following experiments and analyses are conducted with OpenGCL.

\section{Experimental Setup}
\label{sec:setup}
In this section, we describe the modules, hyperparameters, benchmark datasets and design in our experiments. 

\subsection{Modules} We list the ranges for each module in Table~\ref{tab:setup}. To explore the most important model architectures, we consider the module instantiations widely-used in current researches. Note that we name the samplers with their corresponding models. LINE and GAE, DGI and InfoGraph adopt the same samplers, so we denote them as LINE and DGI samplers. We also include MLP and GAT~\cite{velickovic2018graph} as encoders, because they are also widely used in GRL literature.

\subsection{Hyperparameters} 
We select the most essential hyperparameters: embedding size, network layers in the encoder, number of negative samples in the sampler, and learning rate and batch size in the optimization process. The hyperparameter values are chosen from a rigorous range based on experience from current literature. As for hyperparameters in specific modules, like window size in DeepWalk sampler, we fix them at reasonable values. 
We train each model using Adam optimizer~\cite{kingma2014adam} for 500 epochs, and early stop training when there are 3 consecutive loss increase epochs.
\begin{itemize}
    \item Embedding size. We adopt 64 and 128.
    \item Learning rate is fixed at 0.01.
    \item Hidden dimensions. The hidden dimensions are the same as the embedding size.
    \item Encoder layers. The number of encoder layers range from $\{1,2,3,4\}$.
    \item Negative samples. InfoNCE uses in-batch negative samples. JSD uses 1 negative sample for each positive sample.
    \item Sample size. To avoid OOM issues, we sample subgraphs for large graphs. The sample size is 5000.
    \item Batch size. For both node and graph tasks, we constrain the total amount of nodes in a batch. In contrastive learning, large batches are usually considered to be beneficial, so we fix batch size at 4096 nodes. For multi-graph datasets, we select graphs so that the number of all nodes in one batch does not exceed 4096; each graph with more than 4096 nodes (maximum 5000 due to graph sampling) makes up one batch. 
    \item Normalization. Following the common settings, we normalize embeddings for InfoNCE estimator.
\end{itemize}

\begin{table*}[h]
\centering
\caption{Dataset statistics}
\begin{tabular}{llccccc}
\toprule
\textbf{Task} & \textbf{Dataset} & \textbf{\#Graphs} & \textbf{Avg. \#Nodes} & \textbf{Avg. \#Edges} & \textbf{\#Features} & \textbf{\#Classes}\\
\midrule
\multirow{8}{*}{Node Classification} & Amazon-C & 1 & 13,752 & 245,861 & 767 & 10\\
&Amazon-P & 1 & 7,650 & 119,081 & 745 & 8\\
&CiteSeer & 1 & 3,327 & 4,732 & 3,703 & 6\\
&Coauthor-CS & 1 & 18,333 & 81,894 & 6,805 & 15\\
&Coauthor-Phy & 1 & 34,493 & 247,962 & 8,415 & 5\\
&Cora & 1 & 2,708 & 5,429 & 1,433 & 7\\
&PubMed & 1 & 19,717 & 44,338 & 500 & 3\\
&WikiCS & 1 & 11,701 & 216,123 & 300 & 10\\
\midrule 
\multirow{5}{*}{Graph Classification} & IMDB-B & 1,000 & 19.77 & 193.06 & - & 2\\
&IMDB-M & 1,500 & 13.00 & 65.93 & - & 3\\
&PTC-MR & 344 & 14.29 & 14.69 & - & 2\\
&MUTAG & 188 & 17.93 & 19.79 & - & 2\\
&Reddit-B & 2,000 & 508.52 & 497.75 & - & 2\\
\bottomrule
\end{tabular}
\label{tab:dataset}
\end{table*}

\subsection{Datasets} We evaluate the quality of learned node and graph embeddings on downstream classification tasks. For node classification, we adopt eight standard benchmark datasets: Cora, Citeseer, and Pubmed~\cite{sen2008collective} are three widely-used citation networks. Features in Cora and Citeseer are binary word vectors, while in Pubmed, nodes are associated with tf-idf weighted word vectors. Wiki-CS~\cite{mernyei2020wiki} is a newly proposed reference network. Nodes are computer science articles from Wikipedia and edges are hyperlinks linking the articles. Features are the average GloVe~\cite{pennington2014glove} word embeddings for each article and nodes are labeled by their branches in the field. Amazon-Computers and Amazon-Photo~\cite{shchur2018pitfalls} are  built from Amazon co-purchase graph~\cite{mcauley2015image}. Nodes represent products and edges mean two products are frequently bought together. Features of the nodes are bag-of-words vectors of product reviews. Nodes are labeled by product categories. Coauthor-CS and Coauthor-Physics~\cite{shchur2018pitfalls} are co-authorship graphs based on the Microsoft Academic Graph~\cite{sinha2015overview}. Nodes are authors and they are connected if they co-author a paper. Features are bag-of-words vectors of authors' keywords. Nodes are labeled by research fields. We split each dataset into train (20\%), validation (10\%) and test (70\%) data. 

For graph classification, we use another five common datasets: MUTAG and PTC-MR~\cite{chen2007chemdb} are bioinfomatic datasets. MUTAG collects mutagenic aromatic and heteroaromatic
nitro compounds and PTC-MR contains chemical compounds that reports the carcinogenicity for rats. IMDB-B, IMDB-B and IMDB-M~\cite{yanardag2015deep} are collaboration networks for movies. Each graph represent a ego-network for an actor/actress. Nodes are actors/actresses and they are linked if they appear in the same movie. Graphs are labeled by the genres they are generated. 
We use an 80\%/10\%/10\% split for this task. The detailed dataset statistics can be found in Table~\ref{tab:dataset}. Following state-of-the-art approaches~\cite{hassani2020contrastive, sun2020infograph, velickovic2019deep}, we train a logistic regression classifier with five-fold cross-validation.

\subsection{Experiment Design} To quantitatively evaluate the impact of various architectural designs, we conduct \textit{controlled random search} experiments conditioned on module instantiations following~\cite{you2020design}.  For instance, suppose we want to explore the impact of different estimators on a specific task. First we randomly draw $M$ experiments from the full setup space with $I_{\texttt{Est}}=I_{\texttt{InfoNCE}}$. With other modules fixed, we then alter their estimators with $I_{\texttt{Est}}=I_{\texttt{JSD}}$ to get another $M$ experiments, which generates $M$ pairs of experiments for us. To compare InfoNCE with JSD estimator across datasets, we rank experiments inside a pair by their performances and analyze the ranking distribution. Experiments in one pair may rank the same if their performance difference is less than 0.01. In this work, we set $M=100$.
\begin{table*}[ht]
    \centering
    \caption{Search space of modules and hyperparameters.}
    \begin{tabular}{l|l|l}
    \toprule
     & \textbf{Module} & \textbf{Hyperparameter}\\
    \midrule
    Encoder & \tabincell{l}{Node: Lookup, MLP, GCN, GAT, GIN\\ Readout: Mean, Sum} & \tabincell{l}{embedding size,\\network layers}\\
    \midrule
    Sampler & \tabincell{l}{DeepWalk, LINE, DGI, MVGRL, GCA, GraphCL} & negative samples\\
    \midrule
    Discriminator & Inner product, Bilinear &  \\
    \midrule
    Estimator & InfoNCE, JSD & \\
    \midrule
    Optimization &  & learning rate, batch size\\
    \bottomrule
    \end{tabular}
    \label{tab:setup}
\end{table*}

\section{Results and Discussion}
In this section, we discuss the experimental results. To conduct a thorough analysis for over 3,400 experimental runs, we consider three perspectives:
\begin{enumerate}
    \item \textbf{Single module.} Through controlled experiments, we can analyze the impact of each single module. From this perspective, we aim to find how each module would contribute to the final performance. 
    \item \textbf{Pairwise combination.} Apart from single module evaluation, we also explore the effect of pairwise module combination. From this perspective, we try to investigate how the modules interact with each other and find out the best-performing pairs.
    \item \textbf{Full model.} We focus on the assembled model performances on various datasets and compare them with existing models. From this perspective, we want to demonstrate that our framework is potent to produce state-of-the-art models.
\end{enumerate}

\subsection{Single Module Evaluation}
\label{sec:singleg}
\begin{figure*}
\centering
    \begin{subfigure}{\textwidth}
      \centering
      \includegraphics[width=\linewidth]{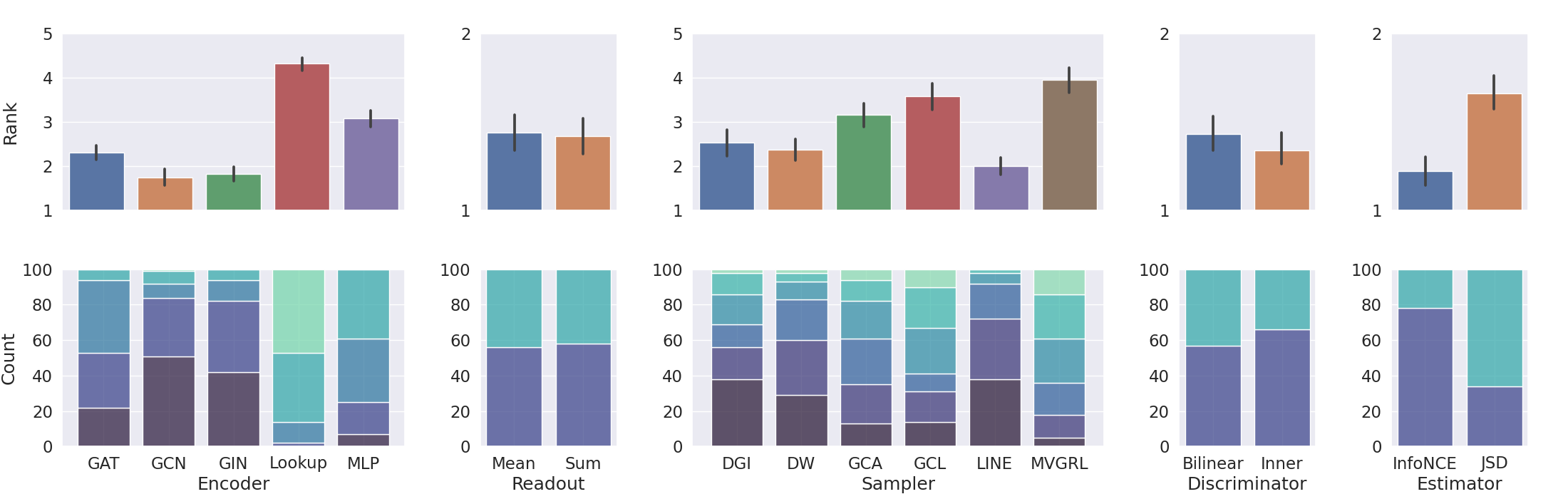}  
      \caption{Node classification.}
      \label{fig:sub-first}
    \end{subfigure}
    \begin{subfigure}{\textwidth}
      \centering
      \includegraphics[width=\linewidth]{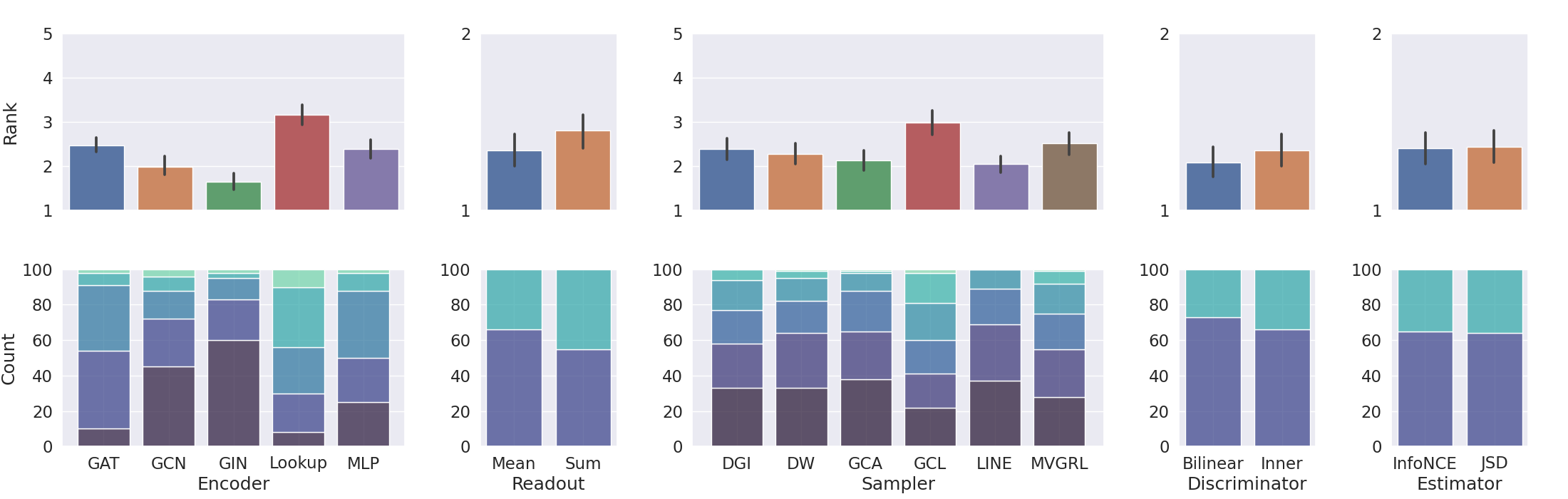}  
      \caption{Graph classification.}
      \label{fig:sub-second}
    \end{subfigure}
        
\caption{Module ranking results. Lower is better. Top: average ranking and confidence, lower is better. Bottom: ranking distribution. Lower rankings are darker. Sums of the same ranks may differ in case of ties.}
\label{fig:single}
\end{figure*}

 We present the ranking results in Figure~\ref{fig:single}. From the results, we have the following observations.

\begin{table}[!h]
\centering
\caption{\textbf{Top}: Memory and time cost of encoders.  \textbf{Bottom}: Time cost of samplers.}
\begin{tabular}{llcccc}
\toprule
 Dataset &  Encoder & Size (KB) &  Time (ms) &  GPU Memory (MB) \\
\midrule
    Cora &     GAT &       869 &       4.34 &           215.45 \\
    &     GCN &       502 &       2.45 &            30.56 \\
    &     GIN &       502 &       3.06 &            46.08 \\
   &  MLP &       501 &       2.25 &            30.56 \\
 &    Lookup &       827 &       1.94 &            15.04 \\
 MUTAG & GAT &  139 &  4.85 &  304.37 \\
  & GCN &       136 &       2.84 &            14.44 \\
&GIN &       137 &       2.91 &            14.54 \\
&MLP &       136 &       2.26 &            14.44 \\
&Lookup &       997 &       2.26 &            14.35 \\
\bottomrule
\end{tabular}

\begin{tabular}{llccc}
\toprule
 Dataset &     Sampler &  Time (ms) & \# Samples &  Time/Sample ($\mu$s) \\
\midrule
    Cora &                    GraphCL &      45.44 &         3686 &                 12.33 \\
 &                    DGI &       9.26 &         4096 &                  2.26 \\
 &                    GCA &      35.72 &         8192 &                  4.36 \\
 &                  MVGRL &    1342.05 &         4096 &                327.65 \\
 &   LINE &     121.72 &        13264 &                  9.18 \\
 &  DW &     578.92 &         2700 &                214.42 \\
   MUTAG &                    GraphCL &     177.98 &         5983 &                 29.75 \\
&                    DGI &      37.61 &         6742 &                  5.58 \\
 &                    GCA &      76.87 &        13484 &                  5.70 \\
 &                  MVGRL &      49.29 &         6742 &                  7.31 \\
 &   LINE &     123.24 &        10813 &                 11.40 \\
 &  DW &     635.45 &         3360 &                189.12 \\
\bottomrule
\end{tabular}
\label{tab:complexity}
\end{table}

\subsubsection{Node classification.}

(1) For node encoders, GNN encoders are generally better than MLP and Lookup encoders. Under most cases, GCN and GIN stay at the top. This demonstrates the effectiveness of GNNs in node classification. In the meantime, GAT is not as robust as GCN and GIN, which aligns previous works (e.g. ~\cite{shchur2018pitfalls}).

(2) For readout functions, Mean and Sum readout functions are roughly as good, which is reasonable because in the node classification task, node representations are not sensitive to the size of the entire graph. 

(3) For samplers, DGI and LINE samplers have the highest probabilities to rank 1st. Additionally, DeepWalk and LINE samplers are surprisingly robust compared to the newly proposed samplers, since they rarely drop out of top 4. 

(4) For discriminators, Inner product discriminator is slightly better than Bilinear. But the difference is marginal.

(5) For estimators, an interesting observation is that InfoNCE estimator outperforms JSD estimator by a large margin, even though the former is rarely used in node classification. This observation aligns with recent research~\cite{chen2020simple} that using more negative samples benefits contrastive representation learning. We highlight this as useful for future research and suggest using InfoNCE rather than JSD for GCL practitioners.

\subsubsection{Graph classification.}

(1) For node encoders, due to the strong expressive power, GIN stands out among encoders and rank 1st in 60\% of experimental runs. However, the advantages of GNN encoders shrink, and MLP even get equal average ranking with GAT. We believe this is mainly caused by the relatively small sizes and simple structures of graphs in this task.

(2) For readout functions and discriminators,
Mean readout function and Bilinear discriminator are slightly better than their alternatives, but the gaps are close. There are no definitive conclusions for the best modules.

(3) For samplers, the ranking results of various samplers are close, and all the average rankings lie within 3. Considering the consistent robustness and simplicity, we can conclude that DeepWalk and LINE samplers are good choices for both tasks. 

\subsubsection{Efficiency Analysis.} 
Apart from performance, we also care about the efficiency of each module, which is essential for real-world deployment and application. In this section, we compare the memory and time cost of different encoders and samplers, the most time-consuming components, to reveal the most and least efficient modules.
To evaluate time and space (CUDA memory) complexity, we test the time and GPU usage of the encoders and samplers on Cora and MUTAG. We select LINE as baseline and then changed encoder or sampler one at a time. For encoders, we consider the size of parameters, the time and CUDA memory cost in forward propagation. For samplers, we count the time and number of samples. The results are listed in Table ~\ref{tab:complexity}; GPU usage of samplers is not listed since we did not use GPU for sampling. Also, since samplers vary on numbers of sample created, we focus on the average time cost per sample.

We can see that all encoders other than the lookup table share a similar size of parameters, but the GAT encoder takes up more than 20 times of GPU memory than the other encoders, along with a significantly slower calculation process. Given that, GAT is the least efficient and memory-saving encoder.

On the other hand, certain samplers behave differently on different graphs. MVGRL is fast on small graphs but dramatically slow on larger graphs, which means that it is hard to perform difussion efficiently on larges graphs. DeepWalk is slow on performing random walks, and it also samples the least nodes among all samplers; 
Considering the amount of sampled nodes, DGI, GCA, and LINE are three most efficient samplers. GCA and LINE sample large numbers of nodes at high speeds, which is beneficial for large-scale graph contrastive learning

\subsection{Pairwise Combination Evaluation}
\label{sec:pair}
\begin{figure*}[!htbp]
\centering

    \includegraphics[width=0.9\linewidth]{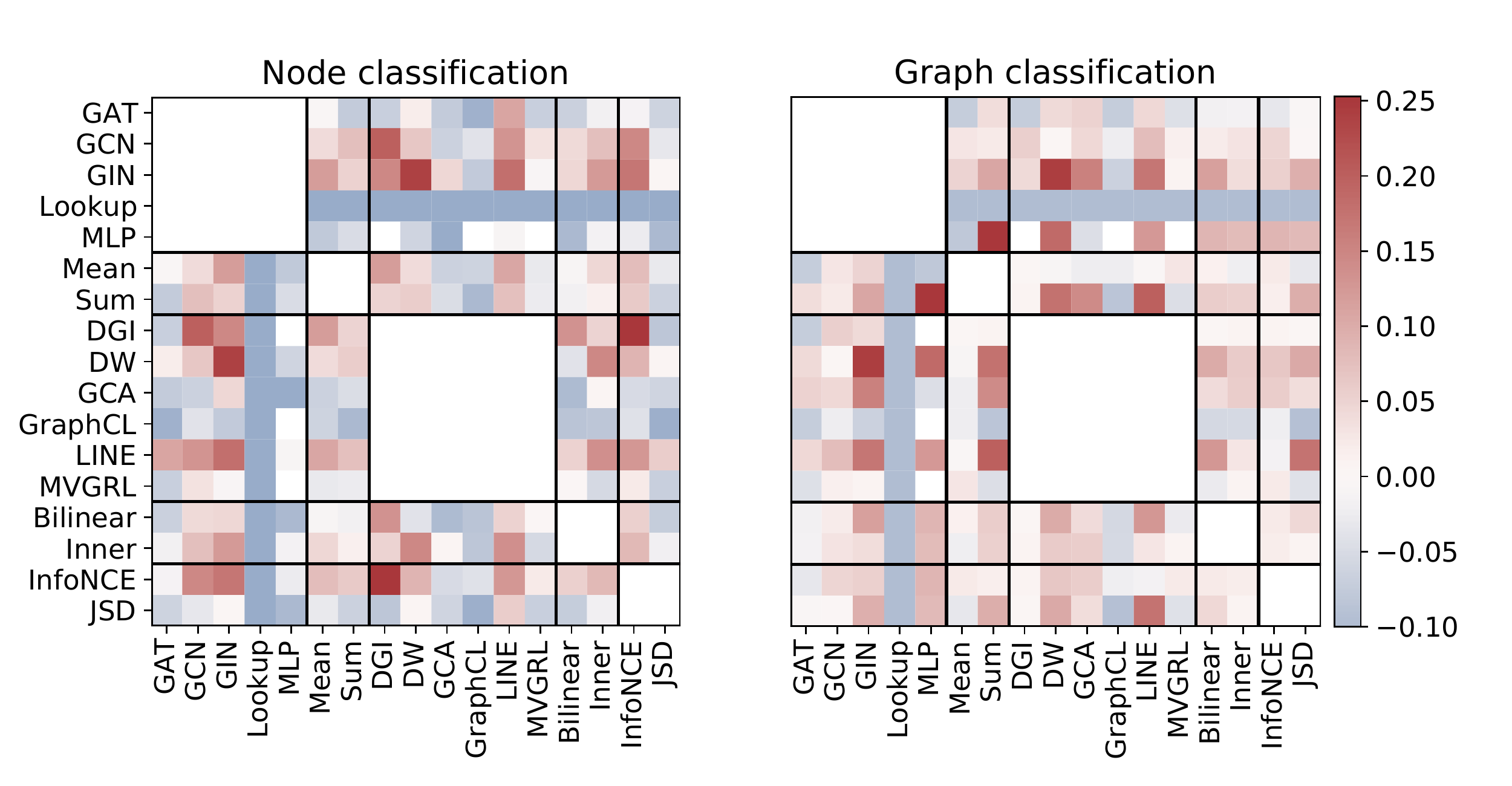}
    \caption{Heatmap of module combination performance $\text{Comb}(x, y)$ for top 10\% ranked runs. Warmer colors in square $(x, y)$ indicate the combination of $x$ and $y$ performs better.}
    \label{fig:pair}
\end{figure*}

To evaluate pairwise combination effect of modules, we group experimental runs by datasets and rank their scores respectively. Then we focus on the top ranked performances of each module combination. Specifically, we collect experimental runs with top $t\%$ scores on each dataset to get a best-$t\%$ pool. For a module pair $(x,y)$, $p_t(x,y)$ denotes the probability for it to enter the best-$t\%$ pool, where
\begin{equation}
    p_t(x,y)=\frac{\# \text{Experimental runs in best-}t\%\ \text{with}\ (x,y) }{\# \text{Experimental runs with}\ (x,y)},
\end{equation}
and we use $\text{Comb}(x, y)=p_t(x,y)-t\%$ to represent the performance of $(x,y)$.
Intuitively, $\text{Comb}(x, y)$ measures the probability difference between experiments with $(x,y)$ and the average.
Figure~\ref{fig:pair} shows the heatmap when $t=10$. The diagonal blocks are masked, since only different kinds of modules can be combined. Our following analysis consists of both macro and micro views: the overall effect of one module on other modules, and the pairwise combination influence of specific modules. The macro view can also be seen as the evaluation of a single module in terms of top ranked experimental runs, which can complement the analysis in previous subsection.

\begin{table*}[!htbp]
\centering
\caption{Best model architecture and baseline on each dataset. The architecture is shown as (Encoder, Readout, Sampler, Discriminator, Estimator).}
\begin{tabular}{lcccc}
\toprule
\textbf{Dataset}      &           \begin{tabular}{@{}c@{}}
                            \textbf{Best sampled architecture}
                    \end{tabular}
                                                     &  \begin{tabular}{@{}c@{}}
                                                            \textbf{Bestscore}
                                                             \end{tabular} 
                                                                            &  \begin{tabular}{@{}c@{}}
                                                                                    \textbf{Best baseline}
                                                                                \end{tabular} 
                                                                                            &  \begin{tabular}{@{}c@{}}
                                                                                            \textbf{Best score of baseline}\end{tabular} \\
\midrule

Amazon-C &    (GCN, Sum, DGI, Bilinear, InfoNCE) &         {\bf0.861} &             GAE &                     0.839 \\
Amazon-P     &  (GIN, Mean, LINE, Bilinear, InfoNCE) &         {\bf0.920} &             GAE &                     0.912 \\
CiteSeer         &       (GIN, Mean, DeepWalk, Inner, InfoNCE) &         {\bf0.708} &             GAE &                     0.680 \\
Coauthor-CS      &             (GAT, Mean, LINE, Inner, JSD) &         {\bf0.940} &             GCA &                     0.931 \\
Coauthor-Phy     &               (GAT, Mean, DeepWalk, Inner, JSD) &         {\bf0.956} &             GCA &                     0.955 \\
Cora             &     (GAT, Mean, LINE, Inner, InfoNCE) &         {\bf0.853} &             GAE &                     0.816 \\
PubMed           &     (GIN, Sum, DeepWalk, Bilinear, InfoNCE) &         {\bf0.831} &             GAE &                     0.830 \\
WikiCS           &      (GAT, Sum, LINE, Inner, InfoNCE) &         {\bf0.773} &             GCA &                     0.762 \\
\midrule
IMDB-B      &    (GCN, Mean, MVGRL, Inner, InfoNCE) &         {\bf0.729} &       InfoGraph &                     0.685 \\
IMDB-M       &     (GIN, Mean, LINE, Inner, InfoNCE) &         {\bf0.530} &       InfoGraph &                     0.483 \\
MUTAG            &            (GCN, Mean, DeepWalk, Bilinear, JSD) &         {\bf0.868} &           MVGRL &                     0.842
 \\
PTC-MR           &             (MLP, Sum, DeepWalk, Bilinear, JSD) &         {\bf0.710} &           MVGRL &                     0.681 \\
Reddit-B    &             (GIN, Sum, DeepWalk, Bilinear, JSD) &         {\bf0.780} &           MVGRL &                     0.729 \\

\bottomrule

\end{tabular}

\label{tab:full}
\end{table*}

\subsubsection{Node classification.} From the macro view, GCN and GIN encoders, LINE sampler and InfoNCE estimator generally incorporate well with other modules and show positive effects to the system. 
From the micro view, GCN works well with DGI sampler and GIN favors DeepWalk sampler. As for estimators and samplers, InfoNCE is especially preferable for DGI sampler. We observe that DGI and MVGRL perform better with Bilinear discriminator, while DeepWalk, GCA and LINE work better with Inner product discriminator. This is reasonable because DGI and MVGRL contrast local and global views, which requires alignment between representations. 

\subsubsection{Graph classification.} From the macro view, readout functions become more important, since graph representations are needed for downstream tasks. Sum readout is generally better than Mean readout for getting high scores. Note that this does not conflict with our previous conclusions in Section~\ref{sec:singleg}, because here we only consider the best-$t\%$ pool. 
From the micro view, there are two obvious beneficial combinations: (1) GIN and DeepWalk sampler. This couple is powerful on both tasks, and we attribute this to the utilization of both neighborhood and random walk messages. (2) MLP and Sum readout. This finding is also remarkable, since it indicates that MLP encoder with Sum readout is comparable with GNN-based encoders on current graph classification task. Mesquita et al.~\cite{mesquita2020rethink} gave a possible explanation that most GNN encoders quickly lead to smooth node representations on small graphs. 

\subsection{Full Model Evaluation}
\label{sec:full}

To highlight that our framework has the potential to find high-performance model architectures, we compare the best models from our experimental runs with the mentioned baseline models, which include LINE, DeepWalk, GAE, DGI, MVGRL and GCA for node classification and DGI, InfoGraph, MVGRL and GraphCL for graph classification. The baseline models are implemented with OpenGCL by assembling their modules.
For baseline models, we exhaustively enumerate the entire hyperparameter space from the same range of previous experiments and select the best ones on each dataset. For a fair comparison, we further sample our experiments to match the number of experimental runs with baseline models.

From Table~\ref{tab:full}, we observe that 
for each dataset, there exists an assembled model that outperforms  the baseline models. The results reveal that searching in the architecture space could be more effective than the hyperparameter space.
Note that due to the differences in dataset splits, tricks and hyperparameters, the baseline results are not exactly the same as those reported in the original papers.

The best model architectures imply that GNN encoders are nearly universal components. It is rather impressive that LINE and DeepWalk samplers occur in the best models on 11 out of 13 datasets. This again supports our conclusion that simple samplers can outperform newly proposed samplers when associated with appropriate modules. 

\section{Conclusion}
Despite the rapid growth of graph contrastive learning models, very few works concentrate on which components are crucial in these models.
In this paper, we evaluate modules in graph contrastive learning by extensive experimental studies, which has been mostly overlooked in previous works. Our results quantify the impact of each single module and pairwise combination. We find that powerful encoders such as GCN and GIN contribute to a substantial fraction of recent progress while samplers should be revisited for their benefits. Furthermore, the full model evaluation shows that our framework has the potential to generate SOTA model architectures.
To foster rigorous comparison methods, we develop OpenGCL as a standard and extensible evaluation toolkit for future research. 

\textbf{Limitations.} As our work mainly focuses on modularized evaluation of current models, we do not pay much attention to hyperparameter tuning, and the selected hyperparameter space is limited. Also, the lack of theoretical analysis prevents us from deeper understandings. In the future, a finer hyperparameter search, more fine-grained submodule-level evaluation, and theoretical analysis may lead to more insightful observations and explanations.

\bibliographystyle{ACM-Reference-Format}
\bibliography{main}

\end{document}